\pdfoutput=1
\documentclass[11pt]{article}
\usepackage[final]{acl} 
\usepackage{times}
\usepackage{latexsym}
\usepackage[T1]{fontenc}
\usepackage[utf8]{inputenc}
\usepackage{microtype}
\usepackage{inconsolata}
\usepackage{graphicx} 
\usepackage{lscape}
\usepackage{multirow}
\usepackage{listings}
\usepackage{pdfpages}
\usepackage{tikz}
\usepackage{booktabs}
\usepackage{amsmath}  
\usepackage{float}
\usepackage[utf8]{inputenc}
\usepackage{makecell}
\usepackage{multirow}
\usepackage{subfigure}
\usepackage{enumitem}
\usepackage{tabularx}
\usepackage{xspace}
\usepackage{adjustbox}
\usepackage{longtable}
\usepackage{listings}
\usepackage{bbding}
\usepackage{afterpage}
\usepackage[utf8]{inputenc}
\usepackage[T1]{fontenc}
\usepackage{CJKutf8}

\title{An Interpretable and Crosslingual Method for Evaluating Second-Language Dialogues} 

 \author{Rena Gao$^\heartsuit$, Jingxuan Wu$^\clubsuit$, Carsten Roever$^\heartsuit$, Xuetong Wu$^\heartsuit$, Jing Wu$^\spadesuit$, Long Lv$^\diamondsuit$, Jey Han Lau$^\heartsuit$ \\     $^\heartsuit$The University of Melbourne, Parkville, 3052, Australia \\
  $^\clubsuit$ The Chinese University of Hong Kong, Shatin, NT, Hong Kong, China\\ 
  $^\diamondsuit$Central South University, Changsha, 410017, China \\
  $^\spadesuit$Tiangong University, Tianjin, 300387, China \\
  \texttt{\{rena.gao,carsten\}@unimelb.edu.au, \{wfyitf,jeyhan.lau\}@gmail.com} \\
  \texttt{wujingxuan@cuhk.edu.cn, wujing@tiangong.edu.cn, lvlong@csu.edu.en}
}

\begin{document}
\maketitle

\begin{abstract}

We analyse the cross-lingual transferability of a dialogue evaluation framework that assesses the relationships between micro-level linguistic features (e.g.\ backchannels) and macro-level interactivity labels (e.g.\ topic management), originally designed for English-as-a-second-language dialogues. To this end, we develop CNIMA (\textbf{C}hinese \textbf{N}on-Native \textbf{I}nteractivity \textbf{M}easurement and \textbf{A}utomation), a Chinese-as-a-second-language labelled dataset with 10K dialogues. We found the evaluation framework to be robust across distinct languages: English and Chinese, revealing language-specific and language-universal relationships between micro-level and macro-level features. Next, we propose an automated, interpretable approach with low data requirement that scores the overall quality of a second-language dialogue based on the framework. Our approach is interpretable in that it reveals the key linguistic and interactivity features that contributed to the overall quality score. As our approach does not require labelled data, it can also be adapted to other languages for second-language dialogue evaluation.
\end{abstract}

\section{Introduction}\label{sec:intro}

In the context of second language (SL) assessment, speaking has always been considered an essential ability  \cite{smith-etal-2022-human,allwood2008dimensions,huang-etal-2020-grade}, but prior studies have predominantly focused on written correction \cite{paiva2022automated} or pronunciation from ASR \cite{mcghee2024highly}. Given the lack of researches that capture the unique linguistic features of SL speakers in dialogues (especially in open-domain interactive conversations), this leaves conversational interaction assessment under-explored and ultimately contributes to a limited understanding of conversational fluency and interactivity for SL speakers from different languages. 

An exception is \citet{gao-etal-2025-interaction}, who propose a two-level framework for assessing the interactivity ability of English-as-a-second-language (ESL) speakers in open-domain conversations. The framework introduces micro-level word/utterance features (e.g.\ backchannels) and macro-level interactivity labels (e.g.\ topic management) and they find that the micro-level features are highly predictive of the macro-level labels. However, they conduct the study only in the context of ESL, raising questions about the transferability of the evaluation framework to other languages besides English.
Furthermore, they do not introduce an automated approach for this evaluation, as the analysis relies on human annotations.  

Our work aims to address these shortcomings by: (1) testing the cross-lingual transferability of the evaluation framework by validating it on an annotated Chinese-as-a-second-language (CSL) dialogue dataset; and (2) introducing an automated, interpretable approach with low data requirement that scores the overall quality of a second-language dialogue. Our automated approach is interpretable as it highlights the key features that lead to the overall score, and it has a low data requirement because it does not require labelled data. As it can be adapted to other languages easily, our method paves the way for a new tool for assessing SL conversations.\footnote{The dataset and related codes can be accessed at \url{https://github.com/RenaGao/CSL2024}} To summarise our contributions:

\begin{itemize}
\item We release CNIMA (\textbf{C}hinese \textbf{N}on-Native \textbf{I}nteractivity \textbf{M}easurement and \textbf{A}utomation), {an annotated} CSL dialogue dataset with 10K dialogues, based on the evaluation framework of \citet{gao-etal-2025-interaction} originally designed for ESL dialogues.
\item  We evaluate the cross-lingual transferability of the evaluation framework and find it to be robust across different languages. We further reveal language-specific and language-universal relationships between micro-level and macro-level features, highlighting the subtle differences between CSL and ESL dialogues.
\item We introduce an automated, interpretable approach that predicts the micro- and macro-level features and the overall quality scores of second-language dialogues. Our best method demonstrates strong performance, creating a new tool for second language assessment. Importantly, the method can be adapted to other languages as it does not require labelled data.
\end{itemize}

\section{Related work}\label{sec:related-work}
\subsection{Assessment of SL Spoken Conversation}  
Mainstream second language assessments in current industrial practice have mainly focused on grammatical accuracy, pronunciation standardization and vocabulary richness; for example, in TOEFL iBT, PTE Academic and Cambridge IELTS test \cite{xu2018measuring,paiva2022automated,xu2021assessing}. However, few speaking assessments emphasised the importance of interaction in dialogues for second language speakers and learners \cite{khabbazbashi2021opening} and aspects such as how speakers manage the topics in communication \cite{shaxobiddin2024discourse}, how speakers perform social roles from speaking interactions \cite{chen2023uncivil}, and how speakers start and end a talk in an acceptable manner \cite{yap2024versatile}.   
There are, however, some exceptions. For example, \citet{dai2022design} develop a test rubric for Chinese Second Language speakers. More recently, \citet{GaoWang+2024} introduce an interactivity scoring framework inspired by the IELTS speaking assessment. However, these studies only provide a theoretical framework for evaluation and lack an automated pipeline for large-scale assessments, human rating is still a major scoring approach in mainstream industrial practice.

\subsection{Automated Scoring System in SL} 

Automated scoring systems can improve the efficiency of processing and scoring dialogues compared to manual assessment methods, allowing for real-time feedback \cite{evanini2017approaches} and explainable suggestions. Some major second language assessment organisations have employed automated speaking scoring, including PTE Pearson \cite{jones2023analyzing}, Duolingo English Test \cite{burstein2021theoretical}, and TOEFL \cite{gong2023challenges}, but these automated systems failed to capture the \textbf{interactive nature} of SL conversations. That is, they struggle to offer detailed analyses and insights on the common errors and language usage patterns and where SL speakers struggle the most. This motivates a more explainable framework for SL assessment that can provide better feedback and suggestions from the automated scoring.  

The main challenge is to design an effective automated evaluation tool. Capturing the nuances of spoken language, such as interaction, attitudes, and cooperation in conversation, is challenging to achieve \cite{GaoWang+2024}. Automated scoring systems must handle diverse patterns across varied languages, which can vary widely among different non-native SL speakers across distinct languages, like Asian languages and European languages with distinct differences. Assessing interactive features like turn-taking, interruptions, and response appropriateness further complicates the process. Ultimately, the dynamic nature of conversations makes it challenging for automated systems to accurately evaluate interactive speaking, necessitating sophisticated algorithms and continuous refinement  \cite{engwall2022identification,cumbal2024robots}.

\section{Evaluation Framework}\label{sec:evaluation-framework}

\begin{figure}
    \centering \includegraphics[width=0.5\textwidth]{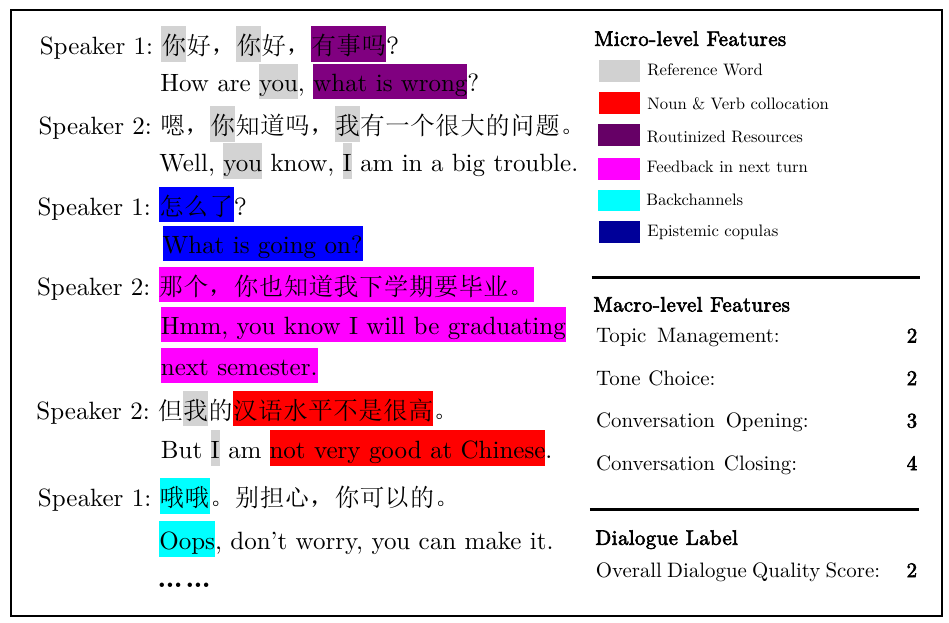} 
    \caption{An example of a CSL dialogue annotated with the micro-level features, macro-level interactivity labels and overall dialogue quality score.} \label{fig:cslexample}
\end{figure}

We extend the evaluation framework of  \citet{gao-etal-2025-interaction} for assessing second-language dialogues, which considers micro-level features and macro-level interactivity labels. One addition we have made is to introduce an overall dialogue quality score. Figure \ref{fig:cslexample} shows an example of CSL dialogue annotation with the evaluation framework. The framework has 17 micro-level features, and it can be further broken into token-level (such as 
 `Reference Word': \textit{she} and `Backchannel': \textit{hmm}) and utterance-level features (such as `Formulaic response': \textit{How's going}). The micro-level features are annotated as spans (i.e.\ annotators mark text spans corresponding to a feature).
 For macro-level interactivity labels, there are four: Topic Management, Tone Choice Appropriateness, Conversation Opening and Conversation Closing, and they are annotated as (dialogue) labels. Briefly, Topic Management refers to the speaker's ability to control and navigate the flow of topics; Tone Choice the suitability of the tone; and Conversation Opening/Closing the naturalness in the initial exchange and conclusion of the discussion. A more elaborate definition can be found in Appendix Table~\ref{tab:dialoguefeaturelabel}. For each macro-level interactivity label, the score ranges from 1 to 5 (categorical), and higher scores indicate a more natural and active interactivity quality. Note that for tone choice appropriateness, higher scores denote a casual tone, and lower scores indicate a formal tone.\footnote{The rationale of assigning higher scores to casual tone is that in our experiments, the conversations are designed to be informal discussions, and appropriate use of casual tone signifies a more natural communicative interaction.}
In addition to micro- and macro-level features, we introduce an \textit{overall score} for each dialogue to measure second language speakers' interaction abilities in open-domain conversation. Like the macro-level labels, it is scored from 1 to 5 (categorical). The overall score is designed to capture the holistic quality of a conversation, integrating elements like contextualisation, responsiveness, and communicative purposes across the whole conversation, and a high score reflects the speaker's ability to engage in fluid and meaningful interactions. The full description of each score is detailed in Appendix Table \ref{tab:score15}.

\section{CNIMA Development}\label{sec:dataset}



\subsection{CSL Dialogue Collection} \label{subsec:CSL-collect}
We first extend the CSL dialogue dataset developed by \citet{wu2021proficiency}, which has approximately 8,000 dialogues.
We followed a similar process, where we recruited 20 learners of Chinese Mandarin at four different proficiency levels.\footnote{Proficiency level can be categorized into beginner, lower, intermediate, and high \cite{wu2021proficiency}.} Each participant was assigned to a dyadic group with a partner of a similar proficiency level. Each pair of participants did one elicited conversation task, in which they discussed an assigned topic, and two role-play activities (three tasks in total; dialogue collection instructions can be found in Appendix~\ref{subapp:SI}).\footnote{All participants have signed consent approved by an ethics board to agree to audio recording.} 
After collecting the conversations, we recruited in-house workers to segment the conversations based on the discussion topics.\footnote{The workers are undergraduate linguistics major students, and they are native Chinese speakers.} Including the original 8,000 dialogues from \citet{wu2021proficiency}, our extended CSL dataset has 10,908 dialogues in total as shown in Table \ref{tab:dataset-statistics}.


\begin{table}[t]
\centering
\begin{tabular}{lll}
\toprule
\toprule
\textbf{Statistic} &  \textbf{Count} \\ 
\midrule
\#dialogues     & 10,908   \\
\#turns (average)   & 6 \\ 
\#turns (max)   & 13 \\ 
\#tokens /w token-level features     & 170,852 \\
\#tokens /w utterance-level features  & 94,516  \\
\bottomrule
\bottomrule
\end{tabular}
\caption{CNIMA Statistics.} \label{tab:dataset-statistics}
\end{table}

\begin{table}[t]
\centering
\resizebox{\columnwidth}{!}{
\begin{tabular}{ccccc}
\toprule
\toprule
\multirow{2}{*}{\textbf{Measure}} &
\textbf{Micro-level} &  \textbf{Macro-level} &\textbf{Overall} \\ 
&\textbf{Features} &  \textbf{Labels} &\textbf{Score}\\
\midrule
 $\alpha$ & {0.66}    & {0.67} & {0.61}  \\ 
 $r$ & {0.65}    &{0.68}  & {0.62}    \\
\bottomrule
\bottomrule
\end{tabular}}
\caption{Inter-annotator agreement for micro-level, macro-level features and overall scores.} \label{tab:annotator-agreement}
\end{table}

\subsection{Data Annotation} \label{subsec:annotate}

Given the dialogues, we annotate them for micro-level features, macro-level interactivity labels and overall quality scores based on the evaluation framework introduced in Section \ref{sec:evaluation-framework}. To this end, we recruited twelve postgraduate students who are native Chinese speakers. We (first author) first trained the annotators and the annotator training manual can be found in Appendix ~\ref{subapp:MN}. During the training, annotators will see one dialogue as an example to understand the requirements and learn how to use the annotation platform (Appendix \ref{subapp:SA}).
After training, each annotator was assigned 950 dialogues, and each dialogue was annotated by two annotators.\footnote{The assignment is created in a way such that we have the same pair of annotators for every batch of 40 dialogues.}
The annotation was conducted over two months, where in the first few weeks we (first author) checked for initial agreement, discussed feedback, and fixed any annotation errors (e.g.\ missing values in the annotation results) before proceeding with the full annotation.
The annotation process was guided by an annotation guide (Appendix~\ref{subapp:MN}), which provided definitions and examples for each micro-level feature, macro-level interactivity label and the overall dialogue quality score. For the overall score, in addition to providing the score, the annotators are also asked to write down their reasons for justifying their label.

As explained in Section \ref{sec:evaluation-framework}, the micro-level features are annotated as spans, and the macro-level labels and overall quality scores as document (dialogue) labels. To aggregate the span annotations by the two annotators for the micro-level features, for each dialogue and each micro-level feature, we iterate through each turn and select the shorter (longer) span between the two annotators if it is a token-level (utterance-level) feature.\footnote{We do this because token-level features (e.g., `Reference word' (She, her, he) or `Tense Choice' (is doing, done, did) tends to be very short and limited to 1 or 2 words and so the shorter spans are likely to be more accurate \cite{greer2023grammar}. The same reasoning applies to utterance-level spans.}
For the macro-level interactivity labels, in dialogues where we have disagreement between the two annotators, we use the \textit{majority label from their larger unsegmented conversation} as the ground truth label.\footnote{Recall that these dialogues are segmented from a larger, longer conversation. The majority label is the most frequent label when we pool together all the labels from the segmented, smaller dialogues that belong to the same original conversation. Our rationale for deferring to this majority label is that these are cases where it is difficult to determine the correct dialogue level label using the smaller segmented dialogue (this happens more with topic management than other interactivity labels), and so we look at all the labels across all the segmented dialogues collection that belong to the same conversation.
}
For the overall dialogue quality scores, for each dialogue with disagreement, we (two authors of this paper) manually assess the justification provided by the annotators to determine the ground truth.\footnote{Note that when resolving the disagreement, we {only} consider the justification \textit{without} looking at their macro-level interactivity labels. This is so we create an unbiased ground truth for the overall score that is independent of the macro-level labels.} 

To measure the annotation quality, for the micro-level features, we calculate agreement between the annotators at the token level for each micro-level feature, i.e., we compute agreement statistics based on the presence or absence of the feature as marked by the annotators for each word token.\footnote{In other words, the unit of analysis here is a word token, and the output is a binary value for each annotator indicating whether it has been marked for the feature.}
We calculate Pearson correlation coefficient $r$ \cite{cohen2009pearson} and Krippendorff's $\alpha$ \cite{krippendorff2018content} and summarise the results in Table~\ref{tab:annotator-agreement}. The agreement is above 0.6 over all the features/labels, indicating a good consensus between annotators. 

\begin{figure}[t]
    \centering  \includegraphics[width=0.4\textwidth]{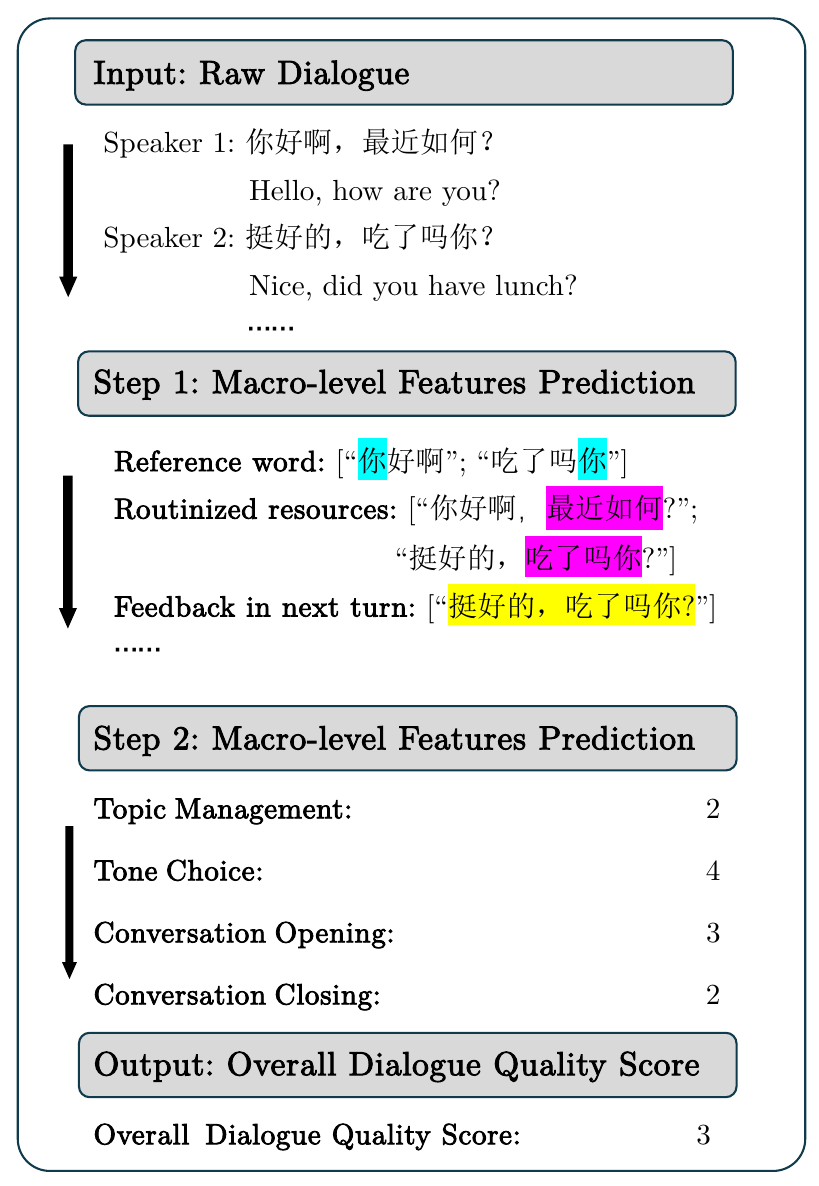}
    \caption{Pipeline for automated scoring of the CSL dialogue on three steps} 
    \label{fig:pipeline}
\end{figure}

\section{Automated Pipeline}

We now propose automating the prediction of the micro-level features (step 1), macro-level interactivity labels (step 2), and overall score (step 3); see Figure~\ref{fig:pipeline} for an illustration. We experiment with classical machine learning models (Logistic Regression (LR), Random Forest (RF), and Naïve Bayes (NB)), fine-tuned Chinese BERT \cite{cui2021pre},
and GPT-4o \cite{OpenAIGPT} with prompts to automate these steps. 
Note that these predictions are done in sequence, where the output from the previous step is used as input for the next step (e.g.\ when predicting the macro-level labels, we use the predicted micro-level features as input).
We partition CNIMA into train, development and test sets (ratio $=$ 7:1:2) for these experiments.

\paragraph{Step 1} This step predicts the spans of the 17 micro-level features given an input dialogue, and we experiment with BERT and GPT-4o here. For BERT, we fine-tune 17 span classifiers (one for each feature) to classify the presence or absence of a feature for each word in the dialogue. For GPT-4o, we do one-shot prompting (i.e.\ 1 dialogue with the expected output as a demonstration) with instructions to generate spans for the token-level and utterance-level features separately.\footnote{In other words, we have 2 prompts for each dialogue.} 


\paragraph{Step 2} This step predicts the macro-level interactivity labels for topic management, tone choice appropriateness, and conversation opening and closing. Here, we experiment with classical models (LR, RF and NB), BERT and GPT-4o. For each of them, we train four (one for each interactivity label) 5-class classifiers (each label has 5 classes). For the classical models, we follow \citet{gao-etal-2025-interaction} and convert the micro-level feature spans into normalised counts and use them as input features to train LR/RF/NB to predict the interactivity labels. For BERT and GPT-4o, the input is the dialogue concatenated with the micro-level feature spans. BERT is fine-tuned as a document classifier, and GPT-4o is one-shot prompted with instructions.


\paragraph{Step 3} This step predicts the overall dialogue quality score. As this is a document classification task like step 2, we follow the same process for building LR, RF, NB, BERT and GPT-4o. Note that the classical models (LR, RF, NB) use only the normalised macro-level interactivity score as input features (in other words, they have only 4 input features), while BERT and GPT-4o uses a concatenation of the dialogue, micro-level feature spans, and macro-level interactivity scores as input.

For more details on configurations and prompts, see Appendix ~\ref{subapp:experimental_details} for BERT and ~\ref{subapp:LLMPro} for GPT-4o.



\section{Experiments}\label{sec:experiment}


We first replicate the ESL experiments of \citet{gao-etal-2025-interaction} for CSL and compare their results in Section \ref{subsec:cross-lingual}. We then assess the performance of our proposed automated approach for SL assessment in Section \ref{subsec:model-analysis}.

\begin{table}[t]
\centering
\begin{tabular}{@{}lcccc@{}}
\toprule
\toprule 
\textbf{Models} & \textbf{Topic} & \textbf{Tone} & \textbf{Opening} & \textbf{Closing} \\
\midrule
LR & 0.829  & 0.859 & 0.821 & 0.810\\
RF & 0.831 & 0.816 & 0.858 & 0.852\\
NB & 0.846 & 0.783 & 0.811 & 0.846 \\
\bottomrule
\bottomrule 
\end{tabular}
\caption{F1 performance of predicting the macro-level features using human-annotated micro-level features.} 
\label{tab:esl-replication-results}
\end{table}

\begin{table*}[t]
\centering
\begin{small}
\begin{tabular}{cccc}
\toprule
\toprule
\textbf{Language} & \textbf{LR} &   \textbf{RF} & \textbf{NB} \\
\midrule
\multirow{5}{*}{{ESL}} & \textbf{Code Switching} & \textbf{Code Switching} & \textbf{Feedback in Next Turn*}\\ 
& \textbf{Reference Word*} & \textbf{Feedback in Next Turn*} & \textbf{Formulaic Responses} \\
& \textbf{Feedback in Next Turn*} & Question-based responses & \textbf{Reference Word*} \\
& \textbf{Formulaic Responses}  & Non-factive Verb & Negotiation of Meaning \\ 
& \textbf{Tense Choice} & \textbf{Reference Word*} & \textbf{Tense Choice} \\ 
\midrule
\multirow{5}{*}{{CSL}} & \textbf{Feedback in the Next Turn} &  \textbf{Subordinate Clauses*} & \textbf{Negotiation of Meaning}\\ 
& \textbf{Noun \& Verb Collocation} &  \textbf{ Routinized Resources} &  \textbf{Noun \& Verb Collocation} \\
& Tense Choice &     \textbf{ Reference Word*} &      \textbf{Routinized Resources} \\
& \textbf{Reference Word*}  &  Code-Switching	&    \textbf{Subordinate Clauses* }\\ 
& \textbf{Subordinate Clauses*} & \textbf{Feedback in Next Turn} &   \textbf{Reference Word*} \\ 
\bottomrule
\bottomrule
\end{tabular}
\end{small}
\caption{High impact common micro-level features over the three classifiers for predicting macro-level features with overlapping features in two/three classifiers by Bold/asterisk. ESL results are reproduced from \citet{gao-etal-2025-interaction}.} \label{tab:common_features}
\end{table*}

\subsection{Transferability of the Evaluation Framework from ESL to CSL} \label{subsec:cross-lingual}
For the \textit{human-annotated} micro-level features, \citet{gao-etal-2025-interaction} convert them into normalised counts and use them as input features to train an LR, NB and RF classifiers to predict macro-level interactivity labels. They found: (1) strong prediction performance; (2) high-impact micro-level features that are \textit{common across all} interactivity labels (by interpreting feature importance given by the trained classifiers); and (3) high-impact micro-level features that are \textit{specifically} predictive for an interactivity label. In the following, we replicate the experiments using our annotated CSL data (CNIMA), to see whether the results transfer across languages; details can be found in Appendix \ref{sec:common-feature-computation}.


For (1), we present the F1 performance of predicting macro-level interactivity labels for CNIMA in Table \ref{tab:esl-replication-results}. We see a strong performance, where F1 is over 0.8 in most models over the 4 interactivity labels. These results echo the ESL results in  \citet{gao-etal-2025-interaction},\footnote{\citet{gao-etal-2025-interaction} found marginally higher performance for conversation opening/closing (F1 scores are over 0.9) and lower performance for topic management and tone choice appropriateness (F1 scores are a little below 0.8).} providing evidence that the evaluation framework is robust across languages.


For (2), we present the results in Table \ref{tab:common_features}. We found that micro-level features such as `Feedback in Next Turn' and `Reference Word' are high-impact features for both ESL and CSL --- this underlines their fundamental role in impacting dialogue interactive dynamics, and it is language-universal. Interestingly, however, we also found some differences. For example, `Noun \& Verb Collocation' and `Routinized Resources' are strong features only for CSL, and this might be because, in Chinese, these fixed terms of expressions are often used to show social closeness in open-domain conversations \cite{roever2021reconceptualizing}.  `Code Switching' and `Tense Choice', on the other hand, are two strong features only for ESL. This is intuitive, as Chinese has no tenses while English tenses correlate with the social expression in communications \cite{lam2018counts}. For the full comparison results, see Appendix Table \ref{tab:response_metrics_bert}.
For (3), Table \ref{tab:csl_labelspefic} and \ref{tab:esl_labelspefic} present the CSL and ESL results, respectively. We found that for topic management, `Negotiation of Meaning' and `Question-Based Responses' are high-impact micro-level features for both ESL and CSL, demonstrating that these are language-universal features important to drive the flow of topics in conversation. For tone appropriateness, we generally see less commonality (exceptions: `Feedback in Next Turn', and `Routinized Resources' which appear in both languages), suggesting that languages tend to use different features for managing tones \cite{zilio2017using}. For conversation opening, we see some similarities (e.g.\ `Question-Based Responses') and also divergences (e.g.\ `Subordinate Clauses' and `Adj./Adv.\ Expressing' for ESL and `Epistemic Copulas' and `Epistemic Modals' for CSL).
One explanation is that English tends to use adjectives and adverbs to extend a topic (e.g.\ \textit{Totally, I think...}), while Chinese prefers modals and copulas \begin{CJK}{UTF8}{gbsn} “应该吧，或许呢”\end{CJK} (translation: \textit{possibly yeah}) to control topic change \cite{alduais2022pragmatic}. 
For conversation closing, we see a more similar trend, where `Collaborative Finishes' and `Backchannels' are strong features across both ESL and CSL. Despite some language-specific variations, the strategies of ending a conversation are largely similar in human communication \cite{lam2021don}.


\begin{table*}[!h]
\centering
\resizebox{2\columnwidth}{!}{
\begin{tabular}{cccc}
\toprule
\textbf{Topic} & \textbf{Tone} & \textbf{Opening} & \textbf{Closing} \\ 
\midrule

\multicolumn{4}{c}{\textbf{Logistic Regression}} \\
\midrule
\textbf{Negotiation of Meaning} & \textbf{Feedback in Next Turn} & \textbf{Epistemic Copulas} & \textbf{Backchannels}\\
\textbf{Epistemic Copulas}  & \textbf{Collaborative finishes}  & Formulaic Responses & \textbf{Collaborative Finishes*}  \\
\textbf{Collaborative Finishes}  & \textbf{Routinized Resources} & \textbf{Question-Based Responses*} &  \textbf{Epistemic Copulas}\\
\textbf{Question-Based Responses} & Formulaic Responses & \textbf{Epistemic Modals} & Subordinate Clauses\\
\textbf{Backchannels*} & \textbf{Question-Based Responses} & Collaborative finishes  &  Formulaic Responses\\
\midrule

\multicolumn{4}{c}{\textbf{Naïve Bayes}} \\
\midrule
Non-factive Verb Phrase  & Epistemic copulas & Code-switching  & Code-switching \\
\textbf{Collaborative Finishes}  & \textbf{Collaborative finishes} & \textbf{Feedback in Next Turn} & Epistemic Modals \\
Code-switching  & Impersonal subject + non-factive verb & \textbf{Epistemic Modals} & \textbf{Epistemic Copulas} \\
\textbf{Backchannels*} & \textbf{Backchannels} &  \textbf{Epistemic Copulas} & \textbf{Collaborative Finishes*} \\
\textbf{Epistemic Copulas} & \textbf{Question-Based Responses} & \textbf{Question-Based Responses*} & \textbf{Question-Based Responses} \\
\midrule

\multicolumn{4}{c}{\textbf{Random Forest}} \\
\midrule
Noun \& verb collocation & Tense choice & \textbf{Feedback in Next Turn} & \textbf{Backchannels} \\
\textbf{Negotiation of Meaning} & \textbf{Backchannels} & Noun \& Verb Collocation & \textbf{Subordinate Clauses}  \\
\textbf{Backchannels*} & Negotiation of Meaning  & Collaborative Finishes  & Formulaic Responses \\
\textbf{Question-Based Responses} & \textbf{Feedback in Next Turn} & \textbf{Question-Based Responses*} & \textbf{Collaborative Finishes*} \\
\textbf{Collaborative Finishes} & \textbf{Routinized Resources} & Formulaic Responses &  \textbf{Question-based Responses}   \\ \midrule
\end{tabular}
}
\caption{CSL: High impact interactivity-specific micro-level features. For each interactivity label, bold/asterisk indicates overlapping features in two/three classifiers.}\label{tab:csl_labelspefic}
\end{table*}

\begin{table*}[!ht]
\centering
\resizebox{2\columnwidth}{!}{
\begin{tabular}{cccc}
\toprule
\textbf{Topic} & \textbf{Tone} & \textbf{Opening} & \textbf{Closing} \\ 
\midrule
\multicolumn{4}{c}{\textbf{Logistic Regression}} \\
\midrule
\textbf{Negotiation of Meaning*} & \textbf{Routinized Resources*} & Epistemic Modals & \textbf{Backchannels*}\\
\textbf{Subordinate Clauses*} & Adj./Adv. Expressing & \textbf{Formulaic Responses} & \textbf{Adj./Adv. Expressing} \\
Noun\&Verb Collocation & \textbf{Feedback in Next Turn*} & \textbf{Question-Based Responses*} & Formulaic Responses \\
\textbf{Question-Based Responses} & \textbf{Formulaic Responses*} & \textbf{Subordinate Clauses*} & \textbf{Collaborative Finishes*} \\
\textbf{Subordinate clauses*} &\textbf{Reference Word}  & \textbf{Adj./Adv. Expressing*} & Epistemic Copulas \\
\midrule

\multicolumn{4}{c}{\textbf{Naïve Bayes}} \\
\midrule
Non-factive Verb Phrase  & \textbf{Routinized Resources*} & \textbf{Adj./Adv. Expressing*} & \textbf{Adj./Adv. Expressing} \\
\textbf{Question-Based Responses} & \textbf{Feedback in Next Turn*} & \textbf{Routinized Resources} & Epistemic Modals \\
Adj./Adv. Expressing & \textbf{Epistemic Copulas} & \textbf{Subordinate Clauses*} & \textbf{Backchannels*} \\
\textbf{Negotiation of Meaning*} & Question-Based Responses & \ Epistemic Copulas  & \textbf{Collaborative Finishes*}  \\
\textbf{Subordinate clauses*} & \textbf{Subordinate Clauses*} & \textbf{Question-Based Responses*} & Question-Based Responses \\
\midrule

\multicolumn{4}{c}{\textbf{Random Forest}} \\
\midrule
\textbf{Negotiation of Meaning*} & \textbf{Epistemic Copulas} & Feedback in Next Turn & Feedback in Next Turn \\
Formulaic Responses & {Backchannels} & \textbf{Subordinate Clauses*} &Subordinate clauses  \\
\textbf{Subordinate Clauses*} & \textbf{Feedback in Next Turn*} & \textbf{Adj./Adv. Expressing*} & \textbf{Collaborative Finishes*} \\
Epistemic Copulas & \textbf{Negotiation of Meaning} & \textbf{Question-Based Responses*} &  \textbf{Formulaic Responses} \\
\textbf{Question-Based Responses} & \textbf{Routinized Resources*} & \textbf{Formulaic Responses} & \textbf{Backchannels*} \\ \midrule
\end{tabular}
}
\caption{ESL: High impact interactivity-specific micro-level features (reproduced from \citet{gao2024interactionmattersevaluationframework}). For each interactivity label, bold/asterisk indicates overlapping features in two/three classifiers.}\label{tab:esl_labelspefic}
\end{table*}

\subsection{Evaluation of Automated Pipeline} \label{subsec:model-analysis}

We now evaluate our automated 3-step approach to predicting the overall dialogue quality score as results in Table~\ref{tab:mainresults}. Our pipeline predicts micro- and macro-level features and dialogue overall quality scores. In terms of model names, each component denotes the model used in a particular step, e.g.\ ``BERT+LR+LR'' means we use BERT for step 1 and LR for steps 2 and 3. Also, ``GPT4'' refers to GPT-4o. 
For brevity, we only include LR results for the classical models as they all have similar performances.
In addition to our 3-step approach, we also include 2 baselines that predict the overall dialogue quality score \textit{directly} based on the input dialogue: (1) fine-tuned BERT (``BERT (One-step)''); and (2)  one-shot GPT-4o with instructions (``GPT4 (One-step)''). We also include a variation where the first step uses human-annotated micro-level features (e.g.\ ``Human+LR+LR'') to understand how much performance degrades when substituting them with predicted features. 

Interestingly, the baselines (``BERT (One-step)'' and ``GPT4 (One-step)'') perform quite poorly, achieving F1 scores of 0.379 and 0.585, respectively. This indicates using the raw dialogue directly for predicting the overall quality is difficult (most studies in NLP, however, follow this setup for dialogue evaluation \cite{finch-etal-2023-dont,zhao-etal-2022-floweval,yang2024structured}). For the variation where we use human-annotated micro-level features (``Human+LR+LR'', ``Human+BERT+BERT'', and ``Human+GPT4+GPT4''), we see that BERT is generally the best model and LR the worst, which is no surprise, given that BERT is pre-trained. GPT4, however, is not far from BERT, even though it is not fine-tuned. Overall, the performance is encouraging, and we see that the best models achieve over 0.80 F1.

When we look at the fully automated pipeline (bottom 4 rows in Table \ref{tab:mainresults}), ``BERT+BERT+BERT'' and ``GPT4+GPT4+GPT4'' perform very strongly (0.807 and 0.791), demonstrating that we have a fully automated system that can reliably assess the quality score of a dialogue. We also notice an interesting trend: When we use either BERT or GPT4 for predicting the micro-level features, LR (i.e.\ ``BERT+LR+LR'' and ``GPT4+LR+LR'') performs very poorly for predicting the overall score, even though the span prediction performance of BERT and GPT4 for the micro-level features is not poor (we will revisit this in Section \ref{sec:additional-analyses}).
This suggests that the classical machine learning models are less tolerant of noise.\footnote{Note that LR does not use the dialogue as input, so any noise in the intermediate features (micro-level or macro-level) will have a bigger impact.}

Taking all these results together, we show that when it comes to assessing SL dialogue quality, it is important to take a pipeline approach to predict important intermediate features (i.e.\ micro- and macro-level) before predicting the overall quality score. This design also has another advantage: it is more interpretable as given an overall score, we can also look at the intermediate results to understand what and where went wrong, providing more meaningful feedback that can benefit both teachers and second language students. Lastly, the strong performance of GPT4 also has another important implication: we can adapt our SL assessment system to another language \textit{without requiring} large-scale manually-annotated data.

\begin{table}[t]
\centering
\begin{small}
\begin{tabular}{@{}lcccc@{}}
\toprule
\toprule 
\textbf{Models} & \textbf{F1} \\
\midrule
BERT (One-step)  & 0.379  \\   
GPT4 (One-step) & 0.585   \\
\midrule
Human+LR+LR & 0.772\\
Human+BERT+BERT & 0.860\\
Human+GPT4+GPT4 & 0.842 \\
\midrule
BERT+LR+LR & 0.329\\
BERT+BERT+BERT  & 0.807   \\
\midrule
GPT4+LR+LR & 0.667 \\
GPT4+GPT4+GPT4 & 0.791 \\
\bottomrule
\bottomrule 
\end{tabular}
\end{small}
\caption{F1 performance of the dialogue overall score for the CSL dialogue in three steps in different models} 
\label{tab:mainresults}
\end{table}

  

\begin{table}[t]
\centering
\begin{small}
\begin{tabular}{@{}lcccc@{}}
\toprule
\toprule 
\textbf{Models} & \textbf{Topic} & \textbf{Tone} & \textbf{Opening} &\textbf{Closing} \\
\midrule
LR  & 0.643& 0.357 & 0.081 & 0.079 \\   
RF  & 0.463 & 0.235 & 0.066 & 0.090  \\
NB  & 0.560 & 0.290 & 0.106 & 0.117  \\
\bottomrule
\bottomrule 
\end{tabular}
\end{small}
\caption{Feature Importance of the four interactivity labels across three machine learning models}
\label{tab:importance_fouraspects}
\end{table}

\subsection{Additional Analyses} 
\label{sec:additional-analyses}

To understand how macro-level interactivity labels contribute to the overall quality score, we present their feature importance as learned by classical machine learning models in the ``Human+LR+LR'', ``Human+RF+RF'' and ``Human+NB+NB'' models in Table~\ref{tab:importance_fouraspects}. Topic management appears to be the most important predictor, followed by tone appropriateness. Conversation opening and closing, on the other hand, has relatively small weights. These observations are supported by \citet{roever2023relationship}, who similarly found that topic flow and tone are dominant in deciding interaction quality.


Next, we assess how well BERT and GPT4 perform for the first (micro-level feature span prediction) and the second-step predictions (macro-level label classification); full results in Appendix\ref{subapp:experimental_details}~Table\ref{tab:response_metrics_bert} and~\ref{tab:response_metrics_gpt} for BERT and GPT4 respectively. To summarise, for most of the 17 micro-level features (step 1), both BERT and GPT4 perform very well (average F1 for BERT and GPT4 is 0.89 and 0.81, respectively). That said, BERT and GPT4 struggle on two features related to non-factive verbs, and this can be attributed to the rarity of these features in the Chinese \cite{roever2023relationship}. Overall, these results show that the proposed micro-features transfer very well from English to Chinese.

For the second step, we look at the performance of ``BERT+BERT'' and ``GPT4+GPT4'' for predicting the macro-level labels (Appendix Table~\ref{tab:communication_metrics}). Across all labels, both BERT and GPT4 perform well, averaging about 0.83 F1 for BERT and 0.78 F1 for GPT4, respectively. Together, these encouraging intermediate evaluation results explain why our pipeline approach can consistently score the overall dialogue quality.

\section{Conclusion} 


We introduce CNIMA, a large-scale annotated dataset for Chinese-as-a-Second-Language (CSL) dialogues based on the evaluation framework introduced by \citet{gao-etal-2025-interaction}. This allows us to test the cross-lingual transferability of the framework, as it is originally designed for English. Our experiments demonstrate that the framework works well for CSL dialogues, suggesting that it is robust across languages and that it is a viable framework for evaluating any second-language conversations. Our analyses also reveal language-specific and language-universal relationships between micro-level linguistic features (such as backchannels) and macro-level interactivity features (such as topic management).
Next, we propose an automated, interpretable approach that predicts the micro- and macro-level features and the overall quality of second-language dialogues. To this end, we explore a range of models, from classical machine learning models to LLMs. Our approach is interpretable because it does not only predict the overall dialogue quality score, but also reveals the key micro-level and macro-level features that contribute to the score. Our LLM-based approach performs very well, and as it does not require any labelled data for training, it can be adapted to other languages easily. Ultimately, our method paves the way for an interpretable and practical system for automatic second-language dialogue assessment.


\section{Limitations}\label{sec:limitations}

The original speech data is transcribed manually into text. Technically we can use speech-to-text software to automate this, but since we are working with second language conversations, this can result in much noise in the translated text. 
The scope of the current paper is limited to CSL, but it would be equally interesting to see how the evaluation framework would work for other SL dialogues, which calls on a joint contribution of the community effort on more SL language dialogue datasets, like Japanese second language datasets, Spanish second language datasets and etc. 


\section*{Ethics Statement}
This study is conducted under the guidance of the ACL Code of Ethics. We manually filtered out potentially offensive content and removed all information related to the identification of annotators.


\bibliography{custom}

\begin{thebibliography}{36}
\expandafter\ifx\csname natexlab\endcsname\relax\def\natexlab#1{#1}\fi

\bibitem[{Alduais et~al.(2022)Alduais, Al-Qaderi, and
  Alfadda}]{alduais2022pragmatic}
Ahmed Alduais, Issa Al-Qaderi, and Hind Alfadda. 2022.
\newblock Pragmatic language development: Analysis of mapping knowledge domains
  on how infants and children become pragmatically competent.
\newblock \emph{Children}, 9(9):1407.

\bibitem[{Allwood(2008)}]{allwood2008dimensions}
Jens Allwood. 2008.
\newblock Dimensions of embodied communication-towards a typology of embodied
  communication.
\newblock \emph{Embodied communication in humans and machines}, pages 257--284.

\bibitem[{Burstein et~al.(2021)Burstein, LaFlair, Kunnan, and von
  Davier}]{burstein2021theoretical}
Jill Burstein, Geoffrey~T LaFlair, Antony~John Kunnan, and Alina~A von Davier.
  2021.
\newblock A theoretical assessment ecosystem for a digital-first
  assessment—the duolingo english test.
\newblock \emph{DRR-21-04}.

\bibitem[{Chen et~al.(2023)Chen, Lau, and Frermann}]{chen2023uncivil}
Ming-Bin Chen, Jey~Han Lau, and Lea Frermann. 2023.
\newblock The uncivil empathy: Investigating the relation between empathy and
  toxicity in online mental health support forums.
\newblock In \emph{Proceedings of the 21st Annual Workshop of the Australasian
  Language Technology Association}, pages 136--147.

\bibitem[{Cohen et~al.(2009)Cohen, Huang, Chen, Benesty, Benesty, Chen, Huang,
  and Cohen}]{cohen2009pearson}
Israel Cohen, Yiteng Huang, Jingdong Chen, Jacob Benesty, Jacob Benesty,
  Jingdong Chen, Yiteng Huang, and Israel Cohen. 2009.
\newblock Pearson correlation coefficient.
\newblock \emph{Noise reduction in speech processing}, pages 1--4.

\bibitem[{Cui et~al.(2021)Cui, Che, Liu, Qin, and Yang}]{cui2021pre}
Yiming Cui, Wanxiang Che, Ting Liu, Bing Qin, and Ziqing Yang. 2021.
\newblock Pre-training with whole word masking for chinese bert.
\newblock \emph{IEEE/ACM Transactions on Audio, Speech, and Language
  Processing}, 29:3504--3514.

\bibitem[{Cumbal(2024)}]{cumbal2024robots}
Ronald Cumbal. 2024.
\newblock \emph{Robots Beyond Borders: The Role of Social Robots in Spoken
  Second Language Practice}.
\newblock Ph.D. thesis, KTH Royal Institute of Technology.

\bibitem[{Dai(2022)}]{dai2022design}
David~Wei Dai. 2022.
\newblock \emph{Design and validation of an L2-Chinese interactional competence
  test}.
\newblock Ph.D. thesis, University of Melbourne (Australia).

\bibitem[{Engwall et~al.(2022)Engwall, Cumbal, Lopes, Ljung, and
  M{\aa}nsson}]{engwall2022identification}
Olov Engwall, Ronald Cumbal, Jos{\'e} Lopes, Mikael Ljung, and Linnea
  M{\aa}nsson. 2022.
\newblock Identification of low-engaged learners in robot-led second language
  conversations with adults.
\newblock \emph{ACM Transactions on Human-Robot Interaction (THRI)},
  11(2):1--33.

\bibitem[{Evanini et~al.(2017)Evanini, Hauck, and
  Hakuta}]{evanini2017approaches}
Keelan Evanini, Maurice~Cogan Hauck, and Kenji Hakuta. 2017.
\newblock Approaches to automated scoring of speaking for k--12 english
  language proficiency assessments.
\newblock \emph{ETS Research Report Series}, 2017(1):1--11.

\bibitem[{Finch et~al.(2023)Finch, Finch, and Choi}]{finch-etal-2023-dont}
Sarah~E. Finch, James~D. Finch, and Jinho~D. Choi. 2023.
\newblock \href {https://doi.org/10.18653/v1/2023.acl-long.839} {Don{'}t forget
  your {ABC}{'}s: Evaluating the state-of-the-art in chat-oriented dialogue
  systems}.
\newblock In \emph{Proceedings of the 61st Annual Meeting of the Association
  for Computational Linguistics (Volume 1: Long Papers)}, pages 15044--15071,
  Toronto, Canada. Association for Computational Linguistics.

\bibitem[{Gao et~al.(2024)Gao, Roever, and
  Lau}]{gao2024interactionmattersevaluationframework}
Rena Gao, Carsten Roever, and Jey~Han Lau. 2024.
\newblock \href {http://arxiv.org/abs/2407.06479} {Interaction matters: An
  evaluation framework for interactive dialogue assessment on english second
  language conversations}.

\bibitem[{Gao et~al.(2025)Gao, Roever, and Lau}]{gao-etal-2025-interaction}
Rena Gao, Carsten Roever, and Jey~Han Lau. 2025.
\newblock \href {https://aclanthology.org/2025.coling-main.729/} {Interaction
  matters: An evaluation framework for interactive dialogue assessment on
  {E}nglish second language conversations}.
\newblock In \emph{Proceedings of the 31st International Conference on
  Computational Linguistics}, pages 10977--11012, Abu Dhabi, UAE. Association
  for Computational Linguistics.

\bibitem[{Gao and Wang(2024)}]{GaoWang+2024}
Wei Gao and Menghan Wang. 2024.
\newblock \href {https://doi.org/doi:10.1515/iral-2023-0258} {Listenership
  always matters: active listening ability in l2 business english paired
  speaking tasks}.
\newblock \emph{International Review of Applied Linguistics in Language
  Teaching}.

\bibitem[{Gong(2023)}]{gong2023challenges}
Kaixuan Gong. 2023.
\newblock Challenges and opportunities for spoken english learning and
  instruction brought by automated speech scoring in large-scale speaking
  tests: a mixed-method investigation into the washback of speechrater in toefl
  ibt.
\newblock \emph{Asian-Pacific Journal of Second and Foreign Language
  Education}, 8(1):25.

\bibitem[{Greer(2023)}]{greer2023grammar}
Tim Greer. 2023.
\newblock Grammar-in-interaction and its place in assessing interactional
  competence.
\newblock \emph{Applied Pragmatics}, 5(2).

\bibitem[{Huang et~al.(2020)Huang, Ye, Qin, Lin, and
  Liang}]{huang-etal-2020-grade}
Lishan Huang, Zheng Ye, Jinghui Qin, Liang Lin, and Xiaodan Liang. 2020.
\newblock \href {https://doi.org/10.18653/v1/2020.emnlp-main.742} {{GRADE}:
  Automatic graph-enhanced coherence metric for evaluating open-domain dialogue
  systems}.
\newblock In \emph{Proceedings of the 2020 Conference on Empirical Methods in
  Natural Language Processing (EMNLP)}, pages 9230--9240, Online. Association
  for Computational Linguistics.

\bibitem[{Jones and Liu(2023)}]{jones2023analyzing}
Jeremy~F Jones and Quanling Liu. 2023.
\newblock Analyzing test-takers’ experiences of high-stakes automated
  language testing.
\newblock \emph{English as a Foreign Language International Journal},
  3(1):1--41.

\bibitem[{Khabbazbashi et~al.(2021)Khabbazbashi, Xu, and
  Galaczi}]{khabbazbashi2021opening}
Nahal Khabbazbashi, Jing Xu, and Evelina~D Galaczi. 2021.
\newblock Opening the black box: Exploring automated speaking evaluation.
\newblock \emph{Challenges in Language Testing Around the World: Insights for
  language test users}, pages 333--343.

\bibitem[{Krippendorff(2018)}]{krippendorff2018content}
Klaus Krippendorff. 2018.
\newblock \emph{Content analysis: An introduction to its methodology}.
\newblock Sage publications.

\bibitem[{Lam(2018)}]{lam2018counts}
Daniel~MK Lam. 2018.
\newblock What counts as “responding”? contingency on previous speaker
  contribution as a feature of interactional competence.
\newblock \emph{Language Testing}, 35(3):377--401.

\bibitem[{Lam(2021)}]{lam2021don}
Daniel~MK Lam. 2021.
\newblock Don’t turn a deaf ear: A case for assessing interactive listening.
\newblock \emph{Applied Linguistics}, 42(4):740--764.

\bibitem[{McGhee et~al.(2024)McGhee, Knill, and Gales}]{mcghee2024highly}
Charles McGhee, Katherine Knill, and Mark Gales. 2024.
\newblock Highly intelligible speaker-independent articulatory synthesis.

\bibitem[{OpenAI(2024)}]{OpenAIGPT}
OpenAI. 2024.
\newblock \href {https://openai.com/index/hello-gpt-4o/} {Hello gpt-4o}.

\bibitem[{Paiva et~al.(2022)Paiva, Leal, and Figueira}]{paiva2022automated}
Jos{\'e}~Carlos Paiva, Jos{\'e}~Paulo Leal, and {\'A}lvaro Figueira. 2022.
\newblock Automated assessment in computer science education: A
  state-of-the-art review.
\newblock \emph{ACM Transactions on Computing Education (TOCE)}, 22(3):1--40.

\bibitem[{Roever and Dai(2021)}]{roever2021reconceptualizing}
Carsten Roever and David~W Dai. 2021.
\newblock Reconceptualizing interactional competence for language testing.
\newblock \emph{Assessing speaking in context: Expanding the construct and its
  applications}, pages 23--49.

\bibitem[{Roever and Ikeda(2023)}]{roever2023relationship}
Carsten Roever and Naoki Ikeda. 2023.
\newblock The relationship between l2 interactional competence and proficiency.
\newblock \emph{Applied Linguistics}, page amad053.

\bibitem[{Shaxobiddin(2024)}]{shaxobiddin2024discourse}
Abdullayev Shaxobiddin. 2024.
\newblock A discourse analysis of modal verbs in modern english: Patterns and
  functions.
\newblock \emph{Journal of new century innovations}, 50(2):145--147.

\bibitem[{Smith et~al.(2022)Smith, Hsu, Qian, Roller, Boureau, and
  Weston}]{smith-etal-2022-human}
Eric Smith, Orion Hsu, Rebecca Qian, Stephen Roller, Y-Lan Boureau, and Jason
  Weston. 2022.
\newblock \href {https://doi.org/10.18653/v1/2022.nlp4convai-1.8} {Human
  evaluation of conversations is an open problem: comparing the sensitivity of
  various methods for evaluating dialogue agents}.
\newblock In \emph{Proceedings of the 4th Workshop on NLP for Conversational
  AI}, pages 77--97, Dublin, Ireland. Association for Computational
  Linguistics.

\bibitem[{Wu and Roever(2021)}]{wu2021proficiency}
Jingxuan Wu and Carsten Roever. 2021.
\newblock Proficiency and preference organization in second language mandarin
  chinese refusals.
\newblock \emph{The Modern Language Journal}, 105(4):897--918.

\bibitem[{Xu(2018)}]{xu2018measuring}
Jing Xu. 2018.
\newblock Measuring “spoken collocational competence” in communicative
  speaking assessment.
\newblock \emph{Language Assessment Quarterly}, 15(3):255--272.

\bibitem[{Xu et~al.(2021)Xu, Jones, Laxton, and Galaczi}]{xu2021assessing}
Jing Xu, Edmund Jones, Victoria Laxton, and Evelina Galaczi. 2021.
\newblock Assessing l2 english speaking using automated scoring technology:
  examining automarker reliability.
\newblock \emph{Assessment in Education: Principles, Policy \& Practice},
  28(4):411--436.

\bibitem[{Yang et~al.(2024)Yang, Zhao, Tang, Zhan, and
  Lin}]{yang2024structured}
Bohao Yang, Kun Zhao, Chen Tang, Liang Zhan, and Chenghua Lin. 2024.
\newblock Structured information matters: Incorporating abstract meaning
  representation into llms for improved open-domain dialogue evaluation.
\newblock \emph{arXiv preprint arXiv:2404.01129}.

\bibitem[{Yap and Sahoo(2024)}]{yap2024versatile}
Foong~Ha Yap and Anindita Sahoo. 2024.
\newblock Versatile copulas and their stance-marking uses in conversational
  odia, an indo-aryan language.
\newblock \emph{Lingua}, 297:103641.

\bibitem[{Zhao et~al.(2022)Zhao, Li, Du, Ji, Yu, Lyu, and
  Wang}]{zhao-etal-2022-floweval}
Jianqiao Zhao, Yanyang Li, Wanyu Du, Yangfeng Ji, Dong Yu, Michael Lyu, and
  Liwei Wang. 2022.
\newblock \href {https://doi.org/10.18653/v1/2022.emnlp-main.715}
  {{F}low{E}val: A consensus-based dialogue evaluation framework using segment
  act flows}.
\newblock In \emph{Proceedings of the 2022 Conference on Empirical Methods in
  Natural Language Processing}, pages 10469--10483, Abu Dhabi, United Arab
  Emirates. Association for Computational Linguistics.

\bibitem[{Zilio et~al.(2017)Zilio, Wilkens, and Fairon}]{zilio2017using}
Leonardo Zilio, Rodrigo Wilkens, and C{\'e}drick Fairon. 2017.
\newblock Using nlp for enhancing second language acquisition.
\newblock In \emph{RANLP}, pages 839--846.

\end{thebibliography}
\newpage
\onecolumn
\appendix

\section{Appendix} \label{sec:appendix}
\subsection{Software Availability} \label{subapp:SA}
To contribute to the research community and facilitate further development and collaboration, we have made the source codes of our innovative annotation tool publicly available. The tool, designed with a focus on enhancing the efficiency and accuracy of data annotation processes, has been developed through meticulous research and development efforts. It incorporates a range of features tailored to meet the needs of researchers and practitioners working in fields that require precise and reliable annotation of datasets.

\subsubsection*{Accessing the Source Code}
The source codes are hosted on GitHub, a platform widely recognized for its robust version control and collaborative features. Interested parties can access the repository at the following link: \url{https://anonymous.4open.science/r/AnnotationTool2023-CFE1/README.md}. This repository is intended for research usage, underlining our commitment to supporting academic and scientific endeavours.

\subsubsection*{Key Features and Capabilities}
Our annotation tool stands out for its user-friendly interface, which simplifies the annotation process and allows users to work more efficiently. Among its key features are:
\begin{itemize}
    \item \textbf{Customizable Annotation Labels:} Users can add their own set of labels to cater to the specific requirements of their projects.
    \item \textbf{Collaborative Annotation Support:} Facilitating teamwork, the tool allows multiple annotators to work on the same dataset simultaneously, ensuring consistency and reducing the time required for project completion.
    \item \textbf{Annotation History Tracking:} All the annotation history, such as changes made, can be tracked, and any further modifications can be done at any time for the user's convenience.
Export Functionality: Annotated data can be exported in several formats, accommodating further analysis or use in machine learning models.
\end{itemize}

\subsection{Pages View For Span Annotation Website Interface} \label{subapp:PV}
\begin{figure}[H]
\centerline{\includegraphics[width=1\linewidth]{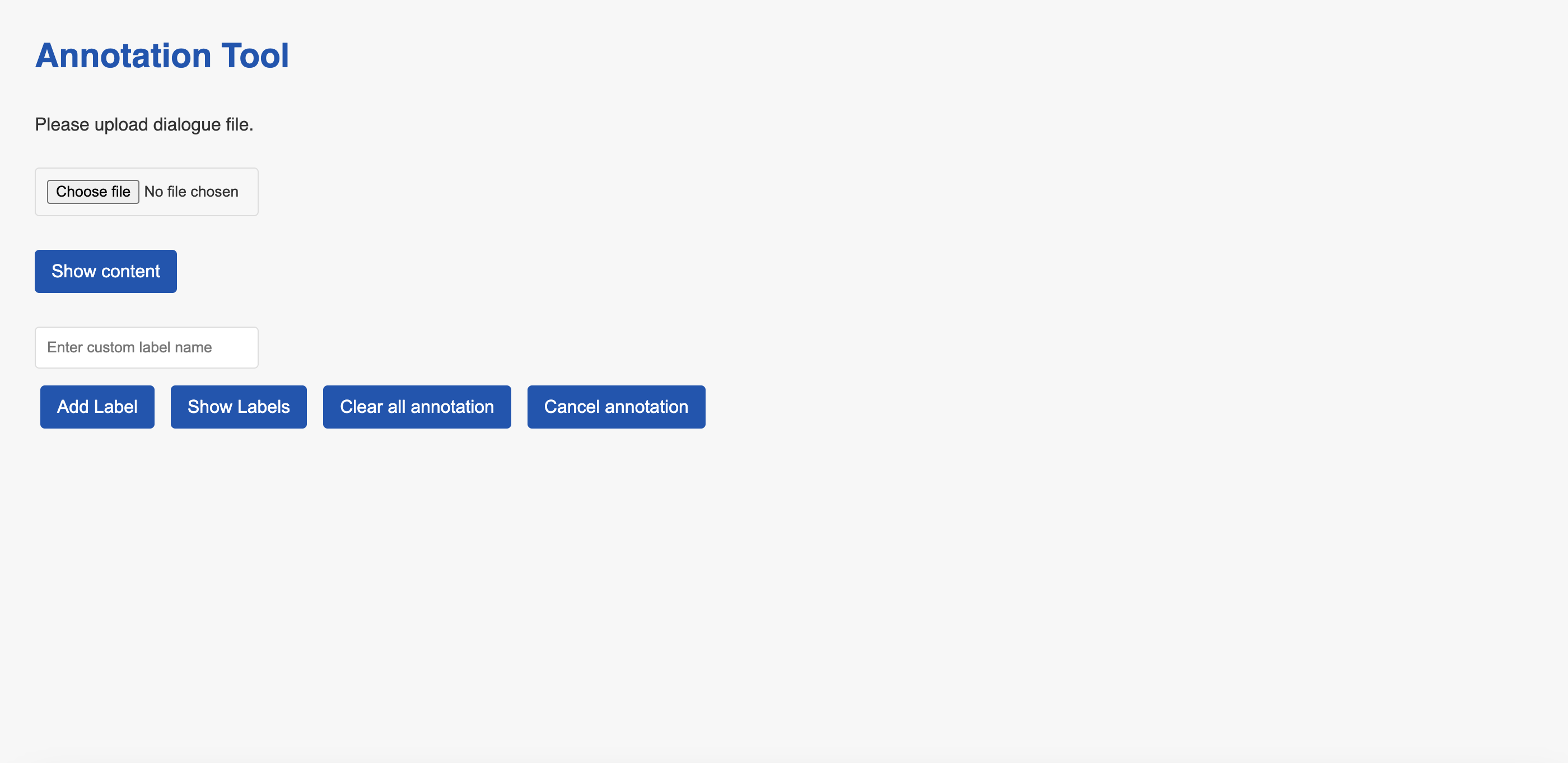}} 
\caption{Annotation tool Demo}
\label{fig:pageview1}
\end{figure}
\begin{figure}[H]
\centerline{\includegraphics[width=1\linewidth]{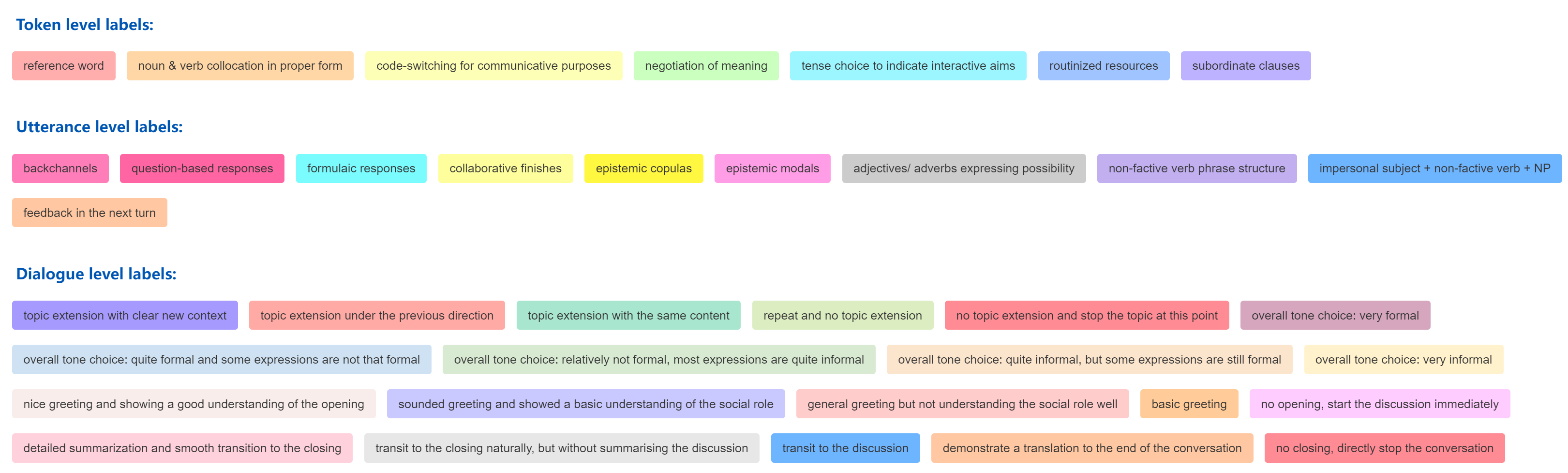}}
\caption{Hierarchical Label Assignment Demo}
\label{fig:pageview2}
\end{figure}




**Token level labels:**

`reference word`　`noun & verb collocation in proper form`　`code-switching for communicative purposes`　`negotiation of meaning`　`tense choice to indicate interactive aims`　`routinized resources`　`subordinate clauses`

**Utterance level labels:**

`backchannels`　`question-based responses`　`formulaic responses`　`collaborative finishes`　`epistemic copulas`　`epistemic modals`　`adjectives/ adverbs expressing possibility`　`non-factive verb phrase structure`　`impersonal subject + non-factive verb + NP`

`feedback in the next turn`

**Dialogue level labels:**

`topic extension with clear new context`　`topic extension under the previous direction`　`topic extension with the same content`　`repeat and no topic extension`　`no topic extension and stop the topic at this point`　`overall tone choice: very formal`

`overall tone choice: quite formal and some expressions are not that formal`　`overall tone choice: relatively not formal, most expressions are quite informal`　`overall tone choice: quite informal, but some expressions are still formal`　`overall tone choice: very informal`

`nice greeting and showing a good understanding of the opening`　`sounded greeting and showed a basic understanding of the social role`　`general greeting but not understanding the social role well`　`basic greeting`　`no opening, start the discussion immediately`

`detailed summarization and smooth transition to the closing`　`transit to the closing naturally, but without summarising the discussion`　`transit to the discussion`　`demonstrate a translation to the end of the conversation`　`no closing, directly stop the conversation`

Figure 4: Hierarchical Label Assignment Demo

## A.3 Manual

### Annotation Munnal for the CNIMA Dataset

**1. Introduction to the annotation task**

The research aims to investigate the interactive ability of second-language speakers of Chinese through automated dialogue evaluation. This study has been approved by **Human Ethics**.

In your annotation, two types of dialogue tasks would be included in this study conducted by a pair-wise discussion by second-language speakers of Chinese participants. The first task is an elicited conversation task, in this part, two speakers will share some experience or what they want to deliver based on the instructions (e.g., *share some ideas on how you think of education in your life*). In the second task, two speakers need to role-play through a joint discussion.

The dialogue of the two tasks was both transcribed into text, and you are ready to annotate based on the text. Please notify the researcher if you find any misaligned information in the transcriptions compared with the original recordings during your annotation.

**2. Hierarchy sequence of the label**

| Label name | Label level | Label tag | example |
|---|---|---|---|
| reference word | Token level labels | [RA] | 你到哪里啦？<br><br>像咱们这样的不是都过了吗？ |
| noun & verb collocation in proper form | | [NVC] | 能帮我个小忙吗？<br><br>帮：动词<br>小忙：名词 |
| code-switching for communicative purposes | | [CS] | 我觉得, well, 这个是吧<br><br>well：英文 |

| | | | |
|---|---|---|---|
| negotiation of meaning (appropriate tense to show meaning) | | [NM] | SPK_1：苏州,苏州<br><br>SPK_2:哦,苏州是南方的城市就比较热一点<br><br>SPK_1：那么看你还是没有做<br><br>SPK_2:的确是的毕竟我一点也没学. |
| tense choice to indicate interactive aims (politeness in talking/ social distance/ context variance) | | [TT] | SPK_1:一两门, 啊, 那我们现在还 能 学 啥 呀？ |
| routinized resources (projector construction) | | [RR] | 哎,你说你 |
| subordinate clauses | | [RC] | 你是数学好<br><br>但是我还是觉得这种东西需要自己用很长的时间弄了,才是 真的会了. |
| backchannels | Utterance level labels | [BC] | 会的, 会有点累 |
| question-based responses | | [QR] | SPK_1:<br><br>我不是把她教的那个知识给反驳了吗？<br><br>SPK_2:<br><br>你既然有反驳的能力,那你还是自学吧. |
| formulaic responses（固定用词） | | [FR] | 但是不管怎么样 我 也 学会了。差不了多少。 |
| collaborative finishes | | [CF] | SPK_1:<br>好的 再见 |

14

| | | | SPK_2：<br>再见 |
|---|---|---|---|
| epistemic copulas | | [H1] | 一个人去还是觉得有点变扭. |
| epistemic modals | | [H2] | 我觉得像苏州<br><br>好像有个这种辅导 |
| adjectives/ adverbs expressing possibility | | [H3] | 我估计可能有些这样的情况 |
| non-factive verb phrase structure | | [H4] | 你可以认识他们<br><br>我姑且能跟上吧 |
| impersonal subject + non-factive verb + NP | | [H5] | 我认为可能他会迟到<br><br>Impersonal Subject (可能): This subject is impersonal because it does not refer to a specific individual but rather expresses a possibility.<br><br>Non-Factive Verb (认为): This verb is non-factive because it introduces a belief or opinion rather than a fact.<br><br>Noun Phrase (他会迟到): This is the noun phrase that completes the sentence, stating what is believed or thought. |
| feedback in the next turn | | [FB] | -你感觉如何？<br><br>-会的, 会 有点 累 |

| | | | |
|---|---|---|---|
| topic extension with clear new context (==change to utterance level, but more information context depends== ) | Dialogue level labels | [T1] | 我只有你这个朋友，你又不肯帮我。别人跟我关系都很一般的啊。你知道我什么意思吧？ |
| topic extension under the previous direction | | [T2] | 说到朋友，我只有你这个朋友。 |
| topic extension with the same content | | [T3] | 你说朋友啊，我觉得吧，很难说。 |
| repeat and no topic extension | | [T4] | 关于朋友的事吗？ |
| no topic extension and stop the topic at this point | | [T5] | 朋友？ |
| conversation opening | | [CO1]<br>[CO2]<br>[CO3]<br>[CO4]<br>[CO5] | CO1: nice greeting and show a good understanding of conversation opening in social interactions.<br><br>CO2: sounded greeting and show a basic understanding of the social role.<br><br>CO3: general greeting and didn't demonstrate a good understanding of the social role.<br><br>CO4: basic greeting.<br><br>CO5: no opening just start the discussion immediately. |
| conversation closing | | [CC1]<br>[CC2]<br>[CC3]<br>[CC4]<br>[CC5] | CC1: detailed summarization and smooth transition to the closing of the conversation. |

|  |  |  | CC2: transit to the closing naturally, but without any summarization of the discussion. |
|---|---|---|---|
|  |  |  | CC3: demonstrate a translation to the end of the conversation. |
|  |  |  | CC4: transit to the end of the discussion. |
|  |  |  | CC5: no closing, just stop the conversation. |
| overall tone choice: very formal |  | [OT1] | 很荣幸与您见面，幸会。 |
| overall tone choice: quite formal and some expressions are not that formal |  | [OT2] | 见到你真好啊，最近如何？ |
| overall tone choice: relatively not formal, most expressions are quite informal |  | [OT3] | 好久不见哎，真是有段日子了啊。 |
| overall tone choice: quite informal, but some expressions are still formal |  | [OT4] | 真是有阵子不见了，别来无恙啊哥们。 |
| overall tone choice: very informal |  | [OT5] | 我天，真是老久没见了铁子，抱一个！ |

3. **Label detailed definitions**

| Label Category | Aspect | Definition |
|---|---|---|
| Reference word | Word choice | A reference word, also known as a referential word or referent, is a linguistic term used to describe a word or expression in a sentence that refers to or stands in place of something else in the text. Reference |

| | | |
|---|---|---|
| | | words are used to avoid repetition and to link different parts of a text together by indicating what a subsequent word or phrase relates to. Reference words can take various forms, including pronouns, demonstratives, and other words that replace or point to nouns or noun phrases. |
| Noun & verb collocation in proper form | | Collocations are words or phrases that habitually occur together, forming a strong and natural linguistic association. In the case of noun-verb collocations, a particular noun is often paired with a particular verb due to convention, tradition, or linguistic patterns. These collocations contribute to the fluency, idiomaticity, and naturalness of language.<br><br>Examples of noun-verb collocations:<br><br>Make a decision: "I need to make a decision."<br>Take a shower: "I usually take a shower in the morning."<br>Catch a cold: "I hope I don't catch a cold."<br>Give a speech: "She gave an inspiring speech." |
| Code-switching for communicative purposes | | Code-switching for communicative purposes refers to the deliberate or subconscious alternation between two or more languages or dialects within a single conversation or utterance by bilingual or multilingual speakers. This linguistic phenomenon is employed to fulfill specific communicative needs or functions, such as clarifying a point, expressing identity, signaling solidarity or |

| | | |
|---|---|---|
| | | distinction, accommodating to the listener's language preference, or conveying concepts and emotions more effectively in one language over another. Code-switching is not merely a random mixing of languages but a sophisticated communicative strategy that reflects the speaker's linguistic competence and cultural awareness, often used to navigate and negotiate the social and contextual dynamics of interaction. |
| Negotiation of meaning (appropriate tense to show meaning) | Contextual tense usage | Negotiation of meaning refers to the interactive process through which speakers of different linguistic backgrounds or competencies collaboratively work to understand each other's intentions, messages, and linguistic expressions when communication breakdowns occur. This involves the use of clarification requests, confirmation checks, comprehension checks, and paraphrasing, among other communicative strategies, to ensure mutual understanding is achieved. The negotiation of meaning is a fundamental aspect of second language acquisition and communicative language teaching, highlighting the dynamic nature of language use and the active role learners play in constructing meaning through interaction. |
| Tense choice to indicate interactive aims (politeness / social distance/ context) | | Tense choice to indicate interactive aims involves the strategic use of verb tenses by speakers to fulfill specific communicative goals or intentions within an interaction. This linguistic strategy encompasses the selection of |

| | | |
|---|---|---|
| | | present, past, future, or perfect tenses to convey nuances of time, mood, or aspect, directly influencing the interpretation and direction of the dialogue. Through careful tense selection, speakers can clarify the timing of events, express certainty or speculation about future occurrences, reflect on past experiences, or emphasize the continuity or completion of actions, all of which serve to enhance the clarity, persuasiveness, or relational dynamics of the communication. Tense choice, therefore, is not merely a grammatical decision but a deliberate tool employed by adept language users to navigate conversations and achieve specific interactive aims. |
| routinized resources (projector construction) | Interactional grammatical device | Routinized resources refer to patterns, practices, or tools that have become standardized and regularly employed within specific contexts or activities. These resources are often developed through repeated use over time, leading to a level of automation or routine in their application. In organizational or social settings, routinized resources help in streamlining processes, reducing the need for decision-making about routine tasks, and ensuring consistency in actions and outcomes. They can include documented procedures, established workflows, habitual practices, or even common language and scripts used in interpersonal interactions. |
| subordinate clauses | | Subordinate clauses, also known as dependent clauses, |

| | | are groups of words that contain a subject and a verb but do not express a complete thought and therefore cannot stand alone as a sentence. They function within a sentence by providing additional information to the main clause, to which they are connected by subordinating conjunctions (such as "because," "although," "when," "if") or relative pronouns (such as "who," "which," "that"). Subordinate clauses serve various roles in sentences, including acting as adjectives, adverbs, or nouns, and are essential for adding complexity, detail, and nuance to communication. Their use enables speakers and writers to articulate relationships of cause and effect, contrast, condition, time, and more, enriching the expressiveness and depth of language. |
|---|---|---|

## A.4  Speaking Task Collection Instruction

**Task One: Elicited Conversation**

*Elicited Conversation Task (ECVA)*

*Instruction:* 在这部分中，你将看到一个话题。请就此话题与你的同伴展开讨论。

*In this section, you will be given a topic. Based on the topic, I would like you to talk together with your partner.*

话题：目前，有很多人喜欢独自旅行，也有一些人喜欢和一组人共同去旅行。请讨论：

Nowadays, some people like traveling independently, while others like travels with a group of persons. Please discuss:

你喜欢自己独自出行旅游还是和一组人一起去旅行？为什么？
Do you like traveling individually or traveling with a group of persons? Why?

*(温馨提示：以下提供一些观点供你参考，在讨论中，你可以根据需要加入自己的新观点。你不必将图中所有的观点全部提及。请你们至少讨论 3 分钟，我会在适当的时候停止你们的讨论。)*
*(Note: Some ideas are provided below. You can use these ideas in your discussions if you want. If you have some new ideas, you can use your own ideas in the discussion. You do not need to cover all the ideas provided below. Please discuss at least 3 minutes. I will stop you if necessary.)*

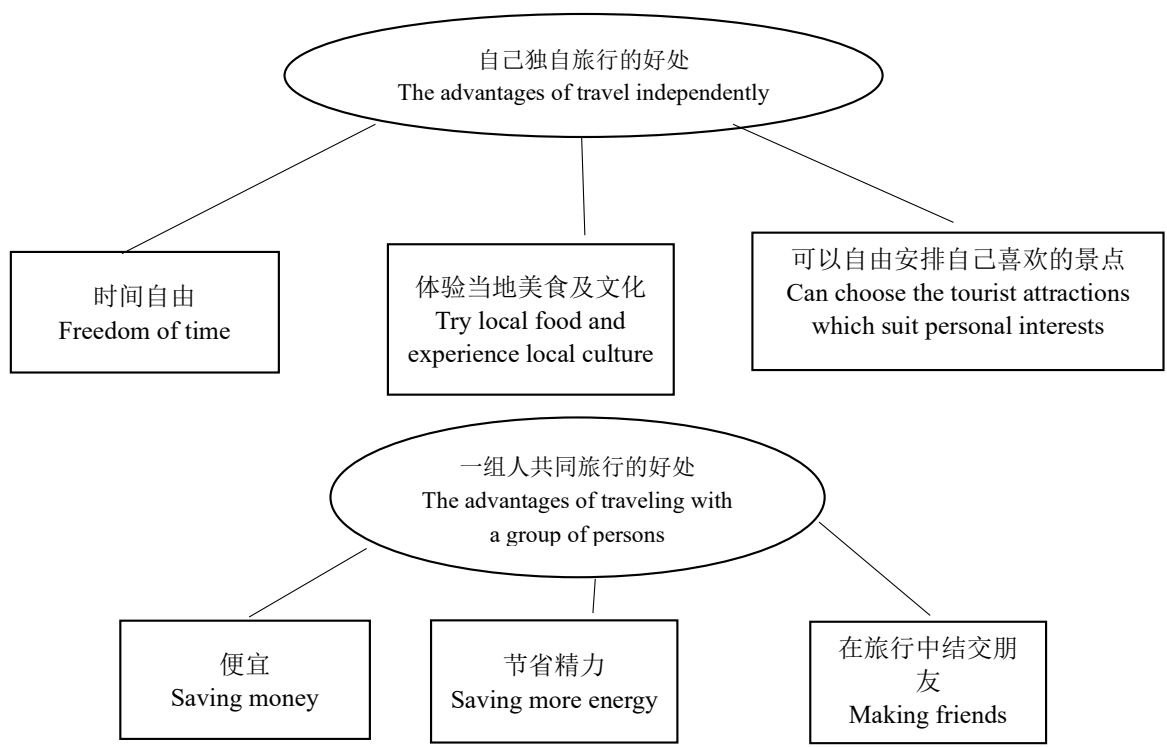

## Task Two: Role Play

**Role Play Task One (VARP1)**

*Role A*

*提示：请你阅读以下场景，想象你自己身处此场景中。如果你有任何问题，请提出。*

*Note: Please read the following situation carefully and imagine yourself in the following scenario. If you have any questions, please feel free to ask.*

你这个学期选择了中国文化课，这门课程要求你使用中文写一篇论文，但是你对自己的中文写作并不是那么的自信。你的好朋友中文水平非常高。你和你的朋友住在同一栋公寓。

现在，你想要去你朋友的房间请求他/她帮你改论文，你会怎么做？

You are enrolled in the Chinese culture course for this semester. One assessment task of this course is to write an essay in Chinese. However, you are not confident in your Chinese writing. Your friend's Chinese proficiency is very high. You and your friend live in the same apartment.

Now, you decide to go to your friend's room. How would you ask your friend to help you revise your essay?

*Role B*

*提示*：

*请你阅读下述场景，你需要想象自己身处这个场景中。你需要以灵活自然的方式引导整个对话。在你的同伴提出请求之前，请不要接受或拒绝你同伴的请求。如果你有任何问题，请提出。*

*Note:*

*Please read the following scenario and imagine yourself in the following situation. It is your responsibility to respond to the task in a natural and flexible way. Before your interlocutor is on the record, please do not accept/ refuse your friends request. If you have any questions, please feel free to ask.*

你这周需要写很多篇论文，下周就是提交这些论文的截止日期了。因此，你非常忙，并且没有足够的睡觉休息时间。你的同学们都知道你的中文水平非常好。

这时，和你同住在一栋公寓的朋友来敲你的房门。

You need to finish writing several papers this week, as the deadline for submitting the papers is next week. Therefore, you are very busy and do not have enough sleep hours. Your classmates know your Chinese proficiency is quite good.

Now, your friend, who lives in the same apartment, rings your doorbell.

**Role Play Task Two (VARP2)**
*Role A*

*提示*:
请你阅读以下场景，想象你自己身处此场景中。如果你有任何问题，请提出。

*Note: Please read the following situation carefully and imagine yourself in the following scenario. If you have any questions, please feel free to ask.*

你想要在网上报名参加汉语水平考试（HSK）。报名 HSK 考试需要使用中国银行卡在网上支付考试费用。但是，你没有中国的银行卡。你想要借你同朋友的银行卡来支付报名费。你的这位朋友和你住在同一栋公寓。

现在，你想要去你朋友的房间，请求他借给你银行卡，你会怎么做？

You want to register for HSK examination online. You need to make an online payment by using a Chinese credit card. However, you do not have a Chinese credit card. You want to borrow a credit card from your friend who lives in the same apartment.

Now, you decide to go to your friend's room. How would ask your friend in real life to borrow the credit card?

 

\subsection{BERT Configuration} \label{subapp:experimental_details}
We give detailed experimental settings in the following table. All the experiments can run via a single NVIDIA GeForce RTX 4070 GPU within a reasonable time. 
\begin{table}[h]
    \centering
    \caption{Experimental Details}
    \label{tab:experimental_details}
    \begin{tabular}{p{5cm}|p{7cm}}
    \hline 
    Parameters & Configurations\\
    \hline 
    Max Length & 128 \\
    Tokenizer & Bert Pretrained Tokenizer  \\
    Encoder & BERT-base-uncase \\
    Activation Function & GELU\\
    Batch Size  & 32 \\ 
    Learning Rate & 5e-5 \\ 
    Loss Function & CrossEntropyLoss \\
    Optimizer & Adam \\
    Epoch & 15 \\
    \hline 
    \end{tabular}

\end{table}

\subsection{Prompts for GPT-4o}\label{subapp:LLMPro}
The following Table 10 shows the prompts for annotating micro-level features with GPT-4o.
\begin{small}
\begin{CJK}{UTF8}{gbsn}
\begin{table}[!h]
    \centering
    \begin{tabular}{|p{4cm}|p{11cm}|}
        \hline
        \textbf{Span} & \textbf{Description and Example} \\
        \hline
        Reference Word & A linguistic term used to avoid repetition and link different parts. \newline \textbf{Example:} \textbf{[你]}到哪里去 (Where are [you] going?) \\
        \hline
        Noun \& Verb Collocation & Words or phrases that habitually occur together, forming a strong and natural linguistic association. \newline \textbf{Example:} 能\textbf{[帮我一个忙]}吗? (Can you [do me a favor]?) \\
        \hline
        Code-switching  & The alternation between languages within a single conversation. \newline \textbf{Example:} 我觉得, [\textbf{well}], 这个是对的 (I think, well, this is right.) \\
        \hline
        Negotiation of Meaning & Interactive process where speakers clarify and confirm understanding. \newline \textbf{Example:} SPK1: 那你是没有做吗? SPK2: \textbf{[的确]}, 我一点也没学 (SPK1: So you didn't do it? SPK2: [Indeed], I haven't learned at all.) \\
        \hline
        Tense Choice to Show Interactive Aims & Using verb tenses to fulfil specific communicative goals. \newline \textbf{Example:} 我\textbf{[现在]}还不能, 因为我还有很多工作 (I can't [right now] because I still have a lot of work.) \\
        \hline
        Routinized Resources & Prefabricated linguistic elements are used to manage dialogue interactions efficiently. \newline \textbf{Example:} \textbf{[你说的你]} ([As you said]) \\
        \hline
        Subordinate Clauses & Clauses that provide additional information to the main clause. \newline \textbf{Example:} 你是数学好, \textbf{[但是]} 我是因为需要自己努力 (You are good at math, [but] because I need to work hard myself.) \\
        \hline
        Backchannels & Brief responses from a listener, such as "uh-huh".\newline \textbf{Example:} \textbf{[哎哎哎。]} ([Yep, yep, yep.])\\
        \hline
        Question Responses & Replies in a dialogue that directly or answer a preceding question.\newline \textbf{Example:}  [我还是自学吧] (SPK2: \textbf{[I shall teach myself.]}) \\
        \hline
        Formulaic Responses & Conventional phrases in dialogue to respond in familiar situations. \newline \textbf{Example:} [差不了多少]\textbf{([Roughly the same])}\\
        \hline
        Collaborative Finishes & Instances in a dialogue where one speaker completes another speaker's sentence or thought. \newline \textbf{Example:} SPK1:[好嘞再见啊] SPK2:[再见了您] \newline 
        (\textbf{SPK1: [Alright, See you.]} \textbf{SPK2:[Goodbye.]})\\
        \hline 
        Epistemic Copulas & Phrases that express a speaker's degree of certainty about a statement, often using verbs like "is" or "seems". \newline \textbf{Example:}  [一个人去还是觉得有点变扭] (\textbf{[I felt wired to go there alone]})\\
        \hline
        Epistemic Modals & Modal verbs or phrases that express a speaker's judgment about the possibility, such as "might," "must,". \newline \textbf{Example:} [你应该自己学会这些中文知识的] (\textbf{[You should learn these Chinese by yourself.]})\\
        \hline
        Adjectives \& adverbs of possibility & Adjectives or adverbs to show possibility, like "Possibly". \newline \textbf{Example:} [我也许回去故乡] (\textbf{[I maybe go back to hometown.]})\\
        \hline
        Non-factive Verb Phrase &  Expressions that use verbs to convey statements without asserting them as true; verbs "think," "believe," or "seem." \newline \textbf{Example:} [我姑且能跟上吧] (\textbf{[I can barely follow the progress.]})\\
        \hline
        Impersonal Subject & An impersonal subject (such as "it" or "there") is followed by a non-factive verb and a noun phrase, often express opinions. \newline \textbf{Example:} [这不好说吧] (\textbf{[It's hard to tell.]})\\
        \hline
        Feedback in Next Turn & Using next turn to respond other speaker. \newline\textbf{Example:} [我认为你有道理] (\textbf{[You words make sense.]})\\
        \hline        
    \end{tabular}
    \label{tab:annotationguidelines}
      \caption{LLM Annotation Prompts for CSL Dialogue Span annotation for Mirco-level features}
\end{table}
\end{CJK}
\end{small}

\subsection{Prompts for GPT-4o Dialogue Overall Evaluation}\label{subapp:LLMeval} 
The following Table 11 shows the prompts for dialogue overall quality score with GPT-4o.

\begin{table}[!ht]
\centering
\begin{tabular}{@{}p{4cm}p{10cm}@{}}
\toprule
\textbf{Field} & \textbf{Description} \\ 
\midrule
\textbf{Conversation} & A dialogue of second language Chinese conversation. \\
\hline 
\textbf{Output Fields} &  \textbf{score}: The score of the interactivity of the Chinese second language dialogue (1 to 5). \\ 
&  \textbf{rationale}: The reason why and how the score is made. \\
\hline 
\textbf{Evaluation Criteria} & 
\textbf{5}: Smooth and fluent daily communication, easy and pleasant. \\
& \textbf{4}: Somewhat less fluent communication, but the communication purpose is achieved. \\
& \textbf{3}: Slightly awkward communication, such as not being able to immediately understand the other person's question with hesitation. \\
& \textbf{2}: Overall communication is not fluent and awkward, but some parts can be mutually understood. \\
& \textbf{1}: Unable to accurately achieve the communication purpose, awkward conversation, failed to talk throughout the conversation. \\
\bottomrule
\end{tabular}
\label{tab:evaluation_task}
\caption{LLM Dialogue Overall Dialogue Quality Evaluation Prompts}
\end{table}

\subsection{Overall Score Description and Definitions}\label{subapp:score15}
The following Table 12 shows the descriptions and definitions for dialogue's overall quality score.
\begin{table*}[!h]
\centering
\begin{tabular}{cl}
\hline
\textbf{Scores} & \textbf{Descriptions}  \\ \hline
5 & \begin{tabular}[c]{@{}l@{}}Smooth and fluent daily communication,\\ easy and pleasant through the whole chat\end{tabular} \\
4 & \begin{tabular}[c]{@{}l@{}}Somewhat less fluent communication,\\ but the communication purpose is achieved\end{tabular}   \\
3 & \begin{tabular}[c]{@{}l@{}}Slightly awkward communication in some places,\\ such as not being able to understand the other person’s question\end{tabular} \\
2 &\begin{tabular}[c]{@{}l@{}}Overall communication is not fluent and mostly awkward,\\ but some parts can be mutually understood\end{tabular} \\
1 & \begin{tabular}[c]{@{}l@{}}Unable to accurately achieve the communication purpose,\\ awkward conversation, and failed to talk throughout the conversation.\end{tabular} \\ \hline
\end{tabular}
\caption{Score description for overall dialogue quality} 
\label{tab:score15}
\end{table*}

\subsection{Four Interactivity Aspects Definitions and Descriptions} 
Table~\ref{tab:dialoguefeaturelabel} shows the descriptions and definitions for dialogue's overall quality score.
\begin{table*}[!h]
\centering
\begin{tabular}{ll}
\hline
\textbf{\begin{tabular}[c]{@{}l@{}}Interactivity \\ Macro-level Features\end{tabular}} &
  \textbf{Definition} \\ \hline
Topic Management &
\begin{tabular}[c]{@{}l@{}}the strategies and techniques used \\ to control and navigate the flow of topics \end{tabular} \\
Tone Choice Appropriateness &
  \begin{tabular}[c]{@{}l@{}}the suitability of the tone used in communication, \\ ensuring it aligns with the context, audience, \\ and purpose to convey the intended message \end{tabular} \\
Conversation Opening &
  \begin{tabular}[c]{@{}l@{}}the initial interaction or exchange that begins a dialogue, \\ often setting the tone and context for the dialogue\end{tabular} \\
Conversation Closing &
  \begin{tabular}[c]{@{}l@{}}the process of ending a dialogue or interaction, \\ which involves signaling the conclusion of the discussion, \\ summarizing key points, and often expressing a farewell\end{tabular} \\ \hline
\end{tabular}
\caption{Definitions of macro-level interactivity features, with higher score emphasising on natural, authentic interaction and active engagement in the dialogue}\label{tab:dialoguefeaturelabel}
\end{table*}

\subsection{Top-k feature computation method}  
\label{sec:common-feature-computation}
This is reproduced from \cite{gao2024interactionmattersevaluationframework}. Given that a trained LR, NB and RF classifier all provide weights to indicate the importance of each feature, for each classifier, we first compute \textit{common} micro-level features $f_{\mathrm{c}}$ across the four interactivity labels:
\begin{align*}
f_{\mathrm{c}} = &\mathrm{top5}\big(\mathrm{top10}(f_{\mathrm{topic}}) \cap \mathrm{top10}(f_{\mathrm{tone}})  \\
&\cap \mathrm{top10}(f_{\mathrm{opening}}) \cap \mathrm{top10}(f_{\mathrm{closing}})\big)
\end{align*}
where $\mathrm{topk}$ is a function that returns the best $k$ items given by their weights, $f_{\mathrm{topic}}$ denote the set of micro-level features with their weights for predicting the topic management interactivity label. 

For micro-level features that are specific to each of the marco-level interactivity aspects. To that end, for each classifier we compute interactivity-specific features, e.g., for topic management, as follows:
\begin{equation}
    \mathrm{top10}(f_{\mathrm{topic}}) - f_c
\end{equation}

\section{Experimental Result}\label{subapp:RE}
\begin{table*}[!ht]
\centering
\begin{tabular}{@{}lcccc@{}}
\toprule
\toprule 
\textbf{Response Type} & \textbf{Accuracy} & \textbf{Precision} & \textbf{Recall} & \textbf{F1 Score} \\
\midrule
Epistemic Copulas & 0.997 & 0.929 & 0.843 & 0.884 \\
Formulaic Responses & 0.976 & 0.875 & 0.781 & 0.825 \\
Question-based Responses & 0.986 & 0.834 & 0.591 & 0.892 \\
Non-factive Verb Phrase Structure & 0.999 & 0.000 & 0.000 & 0.000 \\
Impersonal Subject + Non-factive Verb + NP & 0.997 & 0.909 & 0.243 & 0.384 \\
Reference Word & 0.990 & 0.985 & 0.989 & 0.987 \\
Routinized Resources & 0.983 & 0.783 & 0.706 & 0.742 \\
Noun \& Verb Collocation in Proper Form & 0.985 & 0.970 & 0.958 & 0.964 \\
Collaborative Finishes & 0.995 & 0.802 & 0.631 & 0.706 \\
Tense Choice to Indicate Interactive Aims & 0.994 & 0.957 & 0.930 & 0.943 \\
Negotiation of Meaning & 0.991 & 0.849 & 0.713 & 0.775 \\
Code-switching for Communicative Purposes & 0.999 & 0.976 & 0.954 & 0.965 \\
Feedback in the Next Turn & 0.972 & 0.842 & 0.833 & 0.838 \\
Epistemic Modals & 0.997 & 0.909 & 0.945 & 0.926 \\
Backchannels & 0.982 & 0.824 & 0.731 & 0.875 \\
Subordinate Clauses & 0.990 & 0.960 & 0.937 & 0.948 \\
Adv \& Adj Expressing  & 0.964 &	0.863 &	0.831 &	0.847 \\
\bottomrule
\bottomrule 
\end{tabular}
\caption{Predicted performance of micro-level features on fine-tune BERT} 
\label{tab:response_metrics_bert}
\end{table*}

\begin{table}[!ht]
\centering
\begin{tabular}{@{}lcccc@{}}
\toprule
\textbf{Response Type} & \textbf{Accuracy} & \textbf{Precision} & \textbf{Recall} & \textbf{F1 Score} \\
\midrule
Epistemic Copulas & 0.995 & 0.700 & 0.736 & 0.717 \\
Formulaic Responses & 0.951 & 0.559 & 0.579 & 0.669 \\
Question-based Responses & 0.972 & 0.500 & 0.426 & 0.460 \\
Non-factive Verb Phrase Structure & 0.999 & 0.000 & 0.000 & 0.000 \\
Impersonal Subject + Non-factive Verb + NP & 0.998 & 0.000 & 0.000 & 0.000 \\
Reference Word & 0.987 & 0.982 & 0.983 & 0.982\\
Routinized Resources & 0.971 & 0.564 & 0.448 & 0.500 \\
Noun \& Verb Collocation in Proper Form & 0.972 & 0.953 & 0.900 & 0.826 \\
Collaborative Finishes & 0.988 & 0.565 & 0.406 & 0.472 \\
Tense Choice to Indicate Interactive Aims & 0.990 & 0.915 & 0.826 & 0.768 \\
Negotiation of Meaning & 0.971 & 0.560 & 0.462 & 0.506 \\
Code-switching for Communicative Purposes & 0.999 & 1.000 & 0.941 & 0.869 \\
Feedback in the Next Turn & 0.951 & 0.679 & 0.741 & 0.708 \\
Epistemic Modals & 0.995 & 0.973 & 0.770 & 0.860 \\
Backchannels & 0.975 & 0.697 & 0.588 & 0.638 \\
Subordinate Clauses & 0.964 & 0.863 & 0.831 & 0.847 \\
Adv \& Adj Expressing  & 0.901 &	0.731 &	0.706 &	0.847 \\
\bottomrule
\end{tabular}
\caption{F1 performance of micro-level span annotation by GPT-4o}
\label{tab:response_metrics_gpt}
\end{table}

\begin{table*}[!h]
\centering
\begin{tabular}{@{}lcccc@{}}
\toprule
\toprule 
\textbf{Models} & \textbf{Topic} & \textbf{Tone} & \textbf{Opening} & \textbf{Closing} \\
\midrule
BERT (raw dialogue) & 0.414 & 0.401 & 0.414 & 0.379 \\
GPT-4 (raw dialogue) & 0.553 & 0.533 & 0.585 & 0.557 \\
BERT+BERT (on annotated data) & 0.987 & 0.990 & 0.993 & 0.978\\
\textbf{BERT+BERT} (based on BERT predicted micro-level features) & 0.836 & 0.855 & 0.836 & 0.830 \\
GPT-4+GPT-4 (on annotated data) & 0.761 & 0.749 & 0.812 & 0.809\\
\bottomrule
\bottomrule 
\end{tabular} 
\caption{F1 performance of Marco-level four interactivity aspects' score prediction across different model versions} 
\label{tab:communication_metrics}
\end{table*}

\end{document}